\documentclass[10pt,twocolumn,letterpaper]{article}

\usepackage[pagenumbers]{cvpr} 
\usepackage{algorithm}
\usepackage{multirow} 
\usepackage{bbding} 
\usepackage{fontawesome}
\usepackage{multirow}
\usepackage{times}
\usepackage{epsfig}
\usepackage{graphicx}
\usepackage{amsmath}
\usepackage{amssymb}

\usepackage{booktabs}
\usepackage{epigraph}
\usepackage{enumitem}
\usepackage{float}
\usepackage{chngcntr}
\usepackage{soul}
\usepackage{caption}
\usepackage{subcaption}
\usepackage{setspace} 
\usepackage{zref-xr}
\usepackage{colortbl}
\usepackage{capt-of} %
\usepackage{etoolbox} %
\usepackage[accsupp]{axessibility}
\usepackage[dvipsnames]{xcolor}

\definecolor{cvprblue}{rgb}{0.21,0.49,0.74}
\usepackage[pagebackref,breaklinks,colorlinks,citecolor=cvprblue]{hyperref}

\usepackage{fontawesome}
\newcommand{\authorskip}{\hspace{4.8mm}}

\def\paperID{***}
\def\confName{CVPR}
\def\confYear{2025}

\title{\vspace{-0.5cm}Proteus-ID: ID-Consistent and Motion-Coherent Video Customization \\ 
\vspace*{10pt}
\small\url{https://grenoble-zhang.github.io/Proteus-ID/}
}

\author{
Guiyu Zhang\textsuperscript{1} \hspace{2.5mm}
\authorskip Chen Shi\textsuperscript{1} \hspace{2.5mm}
\authorskip Zijian Jiang\textsuperscript{1}  \hspace{2.5mm}
\authorskip Xunzhi Xiang\textsuperscript{2} \hspace{2.5mm}
\\
\vspace{1mm}
\authorskip Jingjing Qian\textsuperscript{1} \hspace{2.5mm}
\authorskip Shaoshuai Shi\textsuperscript{3} \hspace{2.5mm}
\authorskip Li Jiang\textsuperscript{1~\faEnvelopeO} 
\\ 
{
\fontsize{10pt}{10pt}\selectfont
\textsuperscript{1} The Chinese University of Hong Kong, Shenzhen 
\textsuperscript{2} Nanjing University
}\\
{
\fontsize{10pt}{10pt}\selectfont
\textsuperscript{3} Voyager Research, Didi Chuxing
}
}

\begin{document}

\twocolumn[{%
      \renewcommand\twocolumn[1][]{#1}%
      \maketitle
      \begin{figure}[H]
        \vspace{-1cm}
        \hsize=\textwidth
        \centering
        \includegraphics[width=7 in]{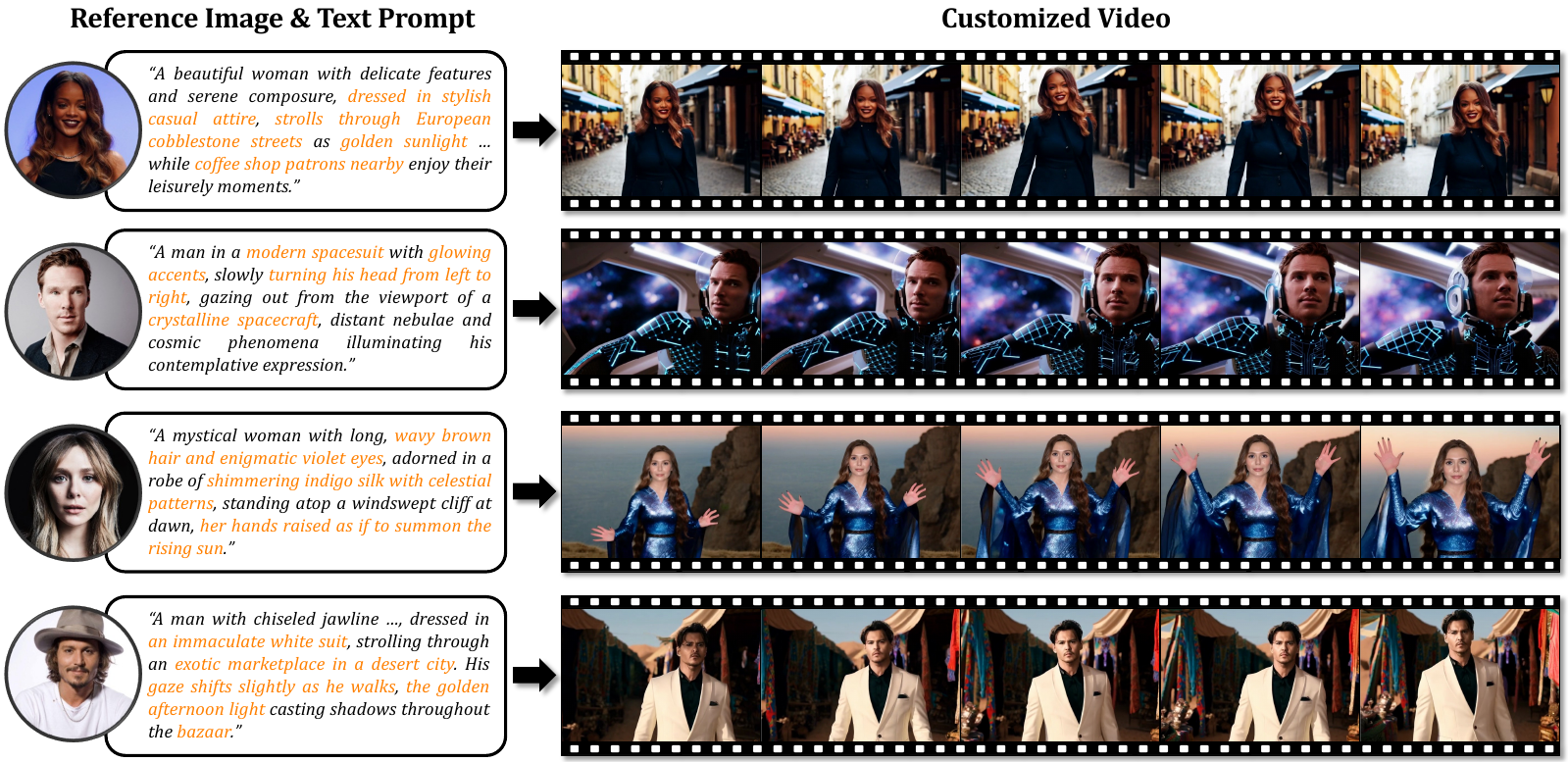}
        \caption{\textbf{Examples of video identity customization generated by Proteus-ID.} Given a reference image and a text description, Proteus-ID can synthesize compelling and expressive animations.
Notably, Proteus-ID effectively preserves identity consistency, maintains semantic alignment across diverse styles, and produces natural, temporally coherent motion. \textcolor{orange}{Orange} highlights attributes mentioned in instructions.}
        \label{fig:teaser}
      \end{figure}
    }]

\renewcommand{\thefootnote}{\fnsymbol{footnote}}
\let\thefootnote\relax\footnotetext{~\faEnvelopeO ~Corresponding author\hspace{5pt}}

\begin{abstract}

Video identity customization seeks to synthesize realistic, temporally coherent videos of a specific subject, given a single reference image and a text prompt. This task presents two core challenges: (1) maintaining identity consistency while aligning with the described appearance and actions, and (2) generating natural, fluid motion without unrealistic stiffness. To address these challenges, we introduce Proteus-ID, a novel diffusion-based framework for identity-consistent and motion-coherent video customization. First, we propose a Multimodal Identity Fusion (MIF) module that unifies visual and textual cues into a joint identity representation using a Q-Former, providing coherent guidance to the diffusion model and eliminating modality imbalance. Second, we present a Time-Aware Identity Injection (TAII) mechanism that dynamically modulates identity conditioning across denoising steps, improving fine-detail reconstruction. Third, we propose Adaptive Motion Learning (AML), a self-supervised strategy that reweights the training loss based on optical-flow-derived motion heatmaps, enhancing motion realism without requiring additional inputs. 
To support this task, 
we construct Proteus-Bench, a high-quality dataset comprising 200K curated clips for training and 150 individuals from diverse professions and ethnicities for evaluation. 
Extensive experiments demonstrate that Proteus-ID outperforms prior methods in identity preservation, text alignment, and motion quality, establishing a new benchmark for video identity customization.

\end{abstract}

\section{Introduction}
Video identity customization aims to synthesize photorealistic videos of a specific subject, given a single reference image and a natural language description~\cite{he2024id, yuan2024identity}. This task requires generating temporally coherent frames that faithfully preserve the subject's identity while performing the actions and appearing in the context described by the prompt. Despite recent progress in text-to-video diffusion models~\cite{kong2024hunyuanvideo, zheng2024open, yang2024cogvideox, zhang2025fantasyid,zhong2025concat}, generating videos that simultaneously maintain identity consistency and follow the semantics of the prompt remains a fundamental challenge.

A major limitation of existing approaches~\cite{zhang2025fantasyid,zhong2025concat} lies in their separate handling of textual prompts and visual references. 
Most models encode the text and image independently and inject them into the denoising model via distinct conditioning pathways, often resulting in semantic conflicts during generation since the denoiser receives misaligned or competing guidance.
This misalignment can cause "copy-paste" artifacts~\cite{yuan2024identity, zhong2025concat, wei2025echovideo},
where the model rigidly replicates the subject’s appearance without integrating the prompt, or loses identity fidelity while following text description.
Such issues stem from the lack of shared semantic grounding between visual and textual identity cues.

To address these limitations, we propose Proteus-ID, a framework for identity-consistent and motion-coherent video generation (see Figure~\ref{fig:teaser}). Built upon prior image-to-video models~\cite{yang2024cogvideox, yuan2024identity}, Proteus-ID introduces three key innovations to enhance generation quality.

First, we introduce a Multimodal Identity Fusion (MIF) module to bridge the semantic gap between visual and textual identity cues. Instead of conditioning the diffusion model on disjoint image and text embeddings, we deeply fuse them into a single joint identity representation. This fusion is achieved via a Q-Former structure that aligns reference image features (captured via both CLIP~\cite{radford2021learning} and face-specific encoders) with an identity description extracted from the prompt. The resulting embedding encapsulates both the subject’s appearance and its semantic interpretation, and is injected into the diffusion model both through the prompt embedding sequence and the cross-attention layers. This unified representation provides coherent and prompt-aligned identity guidance across all frames, significantly mitigating identity-text conflicts.

Second, inspired by prior work on frequency-aware diffusion~\cite{rissanen2022generative, qian2024boosting, yang2024frequency, zhou2025fireedit}, we recognize that diffusion models progressively refine image structure over time, with early timesteps focusing on coarse spatial layout and later ones emphasizing fine-grained details such as facial features. To exploit this property, we propose a Time-Aware Identity Injection (TAII) mechanism that modulates the influence of identity features across timesteps. By integrating timestep embeddings into the fused identity via adaptive transformations, TAII emphasizes coarse traits early and fine-grained details later, ensuring context-aware identity guidance throughout denoising process.


Finally, to improve motion realism—often overlooked in identity-focused generation—we propose an Adaptive Motion Learning (AML) strategy.
Text-to-video models trained with uniform reconstruction loss tend to produce static or rigid outputs, especially when the prompt involves dynamic actions. The proposed AML leverages self-supervised motion signals to reweight the training loss, emphasizing dynamic regions. This guides the model to produce temporally coherent motion without requiring extra input at inference.

To overcome the lack of high-quality training data, we construct Proteus-Bench, a curated dataset to facilitate model training. After meticulous data construction and filtering, we assembled a training dataset of 200K clips.
We evaluate Proteus-ID on 150 individuals spanning diverse professions and ethnicities.
Our method consistently outperforms state-of-the-art baselines in identity similarity, prompt alignment, and motion quality.

Our contributions are as three-fold: (1) We propose a multimodal identity fusion and injection strategy that deeply integrates visual and textual cues to enable identity-consistent and prompt-aligned video generation. (2) We introduce a time-aware identity injection mechanism and an adaptive motion learning scheme that respectively enable dynamic identity conditioning and motion-coherent generation. (3) We construct Proteus-Bench, a new high-quality dataset of  200K curated video clips to support and foster research in video identity customization.

\section{Related Works}
\subsection{Video Identity Customization}
Identity customization aims to preserve distinct identity attributes in generated content. While diffusion models have achieved strong results in identity-preserving image synthesis~\cite{li2024photomaker, wang2024instantid, liang2024caphuman, huang2024consistentid}, extending this to video introduces added complexity—requiring consistency across frames while maintaining natural motion.
Early approaches~\cite{chefer2024still, wang2024customvideo, wei2024dreamvideo, wu2024motionbooth} relied on finetuning a pre-trained model for each new identity during inference, and the high resource consumption limits their practical applicability.
To improve efficiency, recent tuning-free diffusion models have been proposed for video identity customization~\cite{he2024id, polyak2410movie, zhong2025concat, zhang2025fantasyid}. 
Based on AnimateDiff~\cite{guo2023animatediff}, ID-Animator~\cite{he2024id} employs a facial adapter to preserve identity characteristics, but suffers from limitations in text alignment and motion quality. 
The DiT architecture~\cite{zheng2024open, yang2024cogvideox} demonstrates strong potential for improving consistency in video generation. 
Building on the CogVideoX model~\cite{yang2024cogvideox}, ConsisID~\cite{yuan2024identity} and EchoVideo~\cite{wei2025echovideo} use cross-attention to inject facial features and preserve identity. 
FantasyID~\cite{zhang2025fantasyid} incorporates 3D facial geometry to enhance the realism of facial expressions and head poses during video synthesis.
Moreover, Phantom~\cite{liu2025phantom} and VACE~\cite{jiang2025vace} leverage large-scale diverse training data to support multi-subject customization, enhancing the scalability of identity-aware video generation.
Despite these advances, existing methods still suffer from "copy-paste" artifacts and prompt misalignment due to modality conflicts. To address this, we propose Proteus-ID, which introduces a multimodal identity fusion module to jointly encode visual and textual cues—balancing identity preservation with prompt fidelity in video generation.


\subsection{Motion Priors}
\label{related:motion}
Modeling natural, fluid motion from a reference image and text prompt is inherently ambiguous, as a single prompt can correspond to multiple plausible trajectories. 
To address this, many methods incorporate motion priors~\cite{chen2024livephoto, dai2023animateanything, ma2024follow, zhang2024pia, xiang2025remask} to guide motion learning.
Portrait animation approaches often rely on explicit pose or region-based guidance. 
For instance, Animate Anyone~\cite{hu2024animate} adopts a lightweight pose guider to inject pose information into a pre-trained model. MagicAnimate~\cite{xu2024magicanimate} uses DensePose~\cite{guler2018densepose} for motion representation and leverages ControlNet~\cite{zhang2023adding} to encode pose features. Champ~\cite{zhu2024champ} further incorporates SMPL~\cite{loper2023smpl} model sequences, rendered depth, and normal maps to improve motion alignment.
FollowYour-Click~\cite{ma2024follow} leverages optical flow to generate spatial masks for localized motion control. However, these methods depend on auxiliary inputs at training and inference, limiting general usability.
In contrast, we propose an adaptive motion learning strategy that uses optical flow to generate motion heatmaps for loss reweighting—enhancing motion coherence without requiring additional inputs at inference.

\begin{figure*}
	\centering
	\includegraphics[width=0.99\linewidth]{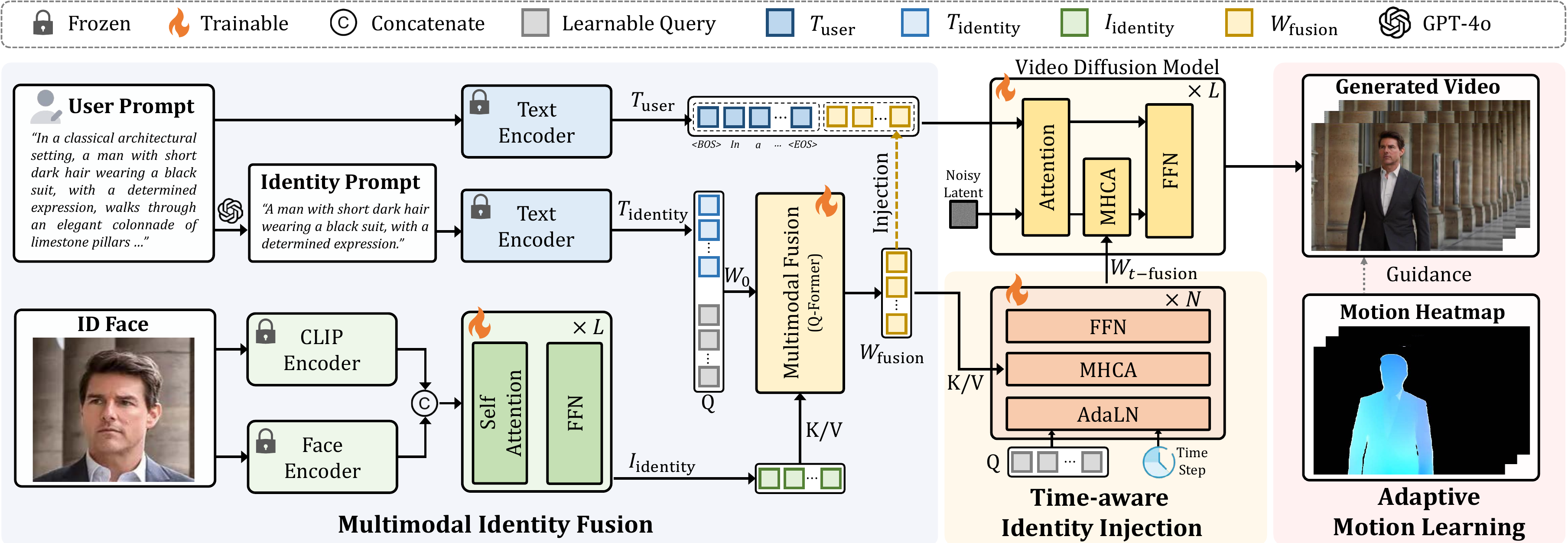}
	\centering
        \vspace{-2mm}
        \caption{\textbf{Overview of Proteus-ID.} Built on a pre-trained DiT, Proteus-ID integrates three key components: Multimodal Identity Fusion (MIF), Time-Aware Identity Injection (TAII), and Adaptive Motion Learning (AML). Given a reference image and user prompt, MIF uses a Q-Former to integrate identity text embeddings with visual features prior to denoising. TAII incorporates timestep embeddings to adaptively modulate identity conditioning during denoising. AML enhances motion realism by introducing a self-supervised motion signal to reweight the training loss—without requiring additional inputs at inference.}
    \label{fig:method}
    \vspace{-3mm}
\end{figure*}

\section{METHOD}
\label{sec:method}
Given a reference image $f$ and a text prompt $y_{\text{user}}$, the video identity customization task aims to generate a realistic, temporally coherent video $\hat{x}$ in which the subject from $f$ preserves their identity while aligning with both the appearance and action described in $y_{\text{user}}$ (see Figure~\ref{fig:method}).
Built upon a diffusion-based video generation framework, our method introduces three key components: (1) a multimodal identity fusion module that unifies visual and textual identity cues (Section~\ref{sec:multimodal}); (2) a time-aware identity injection mechanism that modulates identity influence over denoising steps (Section~\ref{sec:timeaware}); and (3) an adaptive motion learning strategy that improves motion realism through loss reweighting (Section~\ref{sec:motion}). We begin by reviewing the base diffusion model (Section~\ref{sec:pre}).
\subsection{Preliminary}\label{sec:pre}

\noindent
\textbf{Video Diffusion Models.}
Latent diffusion models operate the denoising process in a learned latent space rather than directly in pixel space, greatly improving efficiency and scalability~\cite{ho2020denoising, song2020denoising, rombach2022high}. 
Given a training video $x$, we use a pre-trained variational autoencoder $\varepsilon$ to encode it into a latent tensor $z$, using 3D convolutional layers to capture spatial-temporal structure~\cite{yu2023language}. The forward process adds Gaussian noise over $T$ timesteps to produce noisy latent $z_t$:
\begin{equation}
z_t = \sqrt{\bar{\alpha}_t} z_0 + \sqrt{1 - \bar{\alpha}_t} \epsilon, \quad \epsilon \sim \mathcal{N}(0, I),
\end{equation}
where $\bar{\alpha}_t$ is a fixed noise schedule. The reverse process learns a denoising model $\epsilon_{\theta}$ to predict the added noise. Following~\cite{ho2020denoising}, the basic training objective is
\begin{equation}
L_a \;=\; \mathbb{E}_{t,\;z_0,\;y_{\text{user}},\;\epsilon}\Big[ \big\|\,\epsilon \;-\; \epsilon_{\theta}\!\big(z_t,\, t,\, \tau_{\theta}(y_{\text{user}})\big)\,\big\|^2 \Big], 
\end{equation}
where $\tau_{\theta}(y_{\text{user}})$ is the prompt embedding (e.g., from a T5 encoder).
This trains $\epsilon_{\theta}$ to denoise conditioned on the text description.

\noindent
\textbf{Diffusion Transformer (DiT).}
Recent work has explored transformer architectures for diffusion models in lieu of the traditional UNet backbone~\cite{ronneberger2015u}, which can better capture long-range temporal dependencies and maintain temporal coherence. The Diffusion Transformer (DiT)~\cite{peebles2023scalable} is a latent diffusion model architecture that uses self-attention over spatio-temporal token sequences, improving video coherence and quality. We adopt the MM-DiT from CogVideoX~\cite{yang2024cogvideox} as our base denoising model $\epsilon_{\theta}$.

\subsection{Multimodal Identity Fusion}\label{sec:multimodal}
Prior reference-based video generation methods~\cite{chefer2024still, yuan2024identity, wu2024motionbooth} typically condition diffusion models on visual identity and text separately, often causing conflicts between identity consistency and prompt fidelity. As diffusion models tend to favor visual cues~\cite{luo2022understanding, li2023gligen, huang2023composer}, this imbalance can lead to "copy-paste" artifacts or semantic drift—faithfully replicating the reference face while ignoring the prompt, or vice versa. To address this, we propose to fuse visual and textual identity cues, forming a unified embedding that provides coherent and balanced conditioning to guide the model.

These modalities are naturally complementary: the text conveys general semantics (e.g., “a young woman in glasses”), while the image anchors specific appearance details (e.g., hairstyle, face shape). Conversely, text can refine aspects not visible in the image, such as clothing or action. Fusing both reduces ambiguity and ensures alignment with both identity and prompt semantics.

\noindent
\textbf{Multimodal Identity Encoding.}
To construct a unified identity representation, we first decompose the user prompt $y_{\text{user}}$ into an identity-related phrase $y_{\text{identity}}$ and the remaining action/context description. This is done automatically using a language model (GPT-4o~\cite{achiam2023gpt}). Both $y_{\text{user}}$ and $y_{\text{identity}}$ are encoded using a shared text encoder $\tau_{\theta}$ to obtain text embeddings $T_{\text{user}} = \tau_{\theta}(y_{\text{user}})$ and $T_{\text{identity}} = \tau_{\theta}(y_{\text{identity}})$.

In parallel, we extract visual features from the reference image $f$ using two complementary encoders: a CLIP-based encoder $E_{\text{CLIP}}(f)$ for global appearance, and a face-specific encoder $E_{\text{Face}}(f)$ for fine identity details. These are fused via a transformer network:
\begin{equation}
I_{\text{identity}} = \text{Trans}\big( E_{\text{CLIP}}(f), E_{\text{Face}}(f) \big),
\end{equation}
where $\text{Trans}(\cdot)$ applies attention layers to integrate broad and fine-grained identity cues into single visual embedding. 

\noindent
\textbf{Multimodal Fusion.}
To fuse textual and visual identity cues, we employ Q-Former~\cite{li2022blip} that aligns the identity text embedding with visual features. We initialize its input as:
\begin{equation}
W_0 = [Q, T_{\text{identity}}],
\end{equation}
where $Q$ is a set of learnable query vectors and $[\cdot, \cdot]$ denotes concatenation. The Q-Former processes $W_{0}$ over $L$ transformer layers, each applying multi-head self-attention (MHSA) followed by cross-attention (MHCA) with the visual embedding $I_{\text{identity}}$:
\begin{equation}
\begin{split}
W_l &= \mathrm{FFN}\Big( \mathrm{MHCA}\big( \mathrm{MHSA}(W_{l-1}),\, I_{\text{identity}} \big) \Big), \\
    &\hfill\quad\quad\quad\quad\quad\quad l = 1,\dots,L. \hfill
\end{split}
\end{equation}
This iterative fusion yields $W_{\text{fusion}} = W_L$, a multimodal identity embedding that captures detailed visual appearance from $f$ and semantic traits from $y_{\text{identity}}$. It serves as a unified representation of “this specific person, as described.”

\noindent
\textbf{Injecting Fused Identity into Diffusion.}
To condition the diffusion model on the fused identity, we augment the original text embedding $T_{\text{user}} = \tau_{\theta}(y_{\text{user}})$ with $W_{\text{fusion}}$. Specifically, we project $W_{\text{fusion}}$ to the same dimension as text tokens via a linear layer and append it to the prompt embedding sequence:
\begin{equation}
[T_{\text{user}}, W_{\text{fusion}}] = \psi_{\theta}(y_{\text{user}}, y_{\text{identity}}, f),
\end{equation}
where $\psi_{\theta}$ represents the full multimodal fusion pipeline.

This enriched conditioning improves alignment between identity and prompt semantics. The training objective is:
\begin{equation}
\begin{aligned}
L_b \;=&\; E_{t,\;z_0,\;y_{\text{user}},\;y_{\text{identity}},\;f,\;\epsilon}\Big[ \big\|\, \epsilon \;- \\
&\; \epsilon_{\theta}\!\big(z_t,\, t,\, \psi_{\theta}(y_{\text{user}}, y_{\text{identity}}, f)\big) \,\big\|^2 \Big],
\end{aligned}
\end{equation}
which extends the base loss by explicitly conditioning on both the fused identity and the full prompt. Intuitively, $\epsilon_{\theta}$ learns to denoise while "seeing" a combined representation of who the subject is and what they are doing, allowing it to preserve identity without sacrificing alignment to the text.

\subsection{Time-Aware Identity Injection}\label{sec:timeaware}
Recent studies~\cite{rissanen2022generative, qian2024boosting, yang2024frequency, zhou2025fireedit} show that diffusion models process information unevenly across timesteps: early steps capture coarse, low-frequency structure, while later ones refine high-frequency details. For identity guidance, this implies a need to shift from semantic cues early on to fine-grained features later. Applying the fused identity embedding $W_{\text{fusion}}$ uniformly across timesteps can be suboptimal—too much early guidance may suppress prompt structure, while excessive late guidance can blur fine details.

To address this, we propose a Time-Aware Identity Injection (TAII) module that adaptively modulates identity conditioning over diffusion steps. Inspired by the Resampler~\cite{alayrac2022flamingo}, TAII transforms $W_{\text{fusion}}$ into a timestep-dependent embedding as:
\begin{equation}
W_{t\text{-fusion}} = \phi_{\theta}(W_{\text{fusion}}, t),
\end{equation}
using $N$ stacked blocks that incorporate timestep embeddings via adaptive normalization, cross-attention, and feed-forward layers. This enables $W_{t\text{-fusion}}$ to emphasize coarse identity traits for large $t$ and fine details for small $t$.

We integrate $W_{t\text{-fusion}}$ into the diffusion model by updating each MM-DiT transformer block to cross-attend to it:
\begin{equation}
Z'_i \;=\; Z_i \;+\; \mathrm{MHCA}\!\Big( Q = Z_i,\; K = W_{t\text{-fusion}},\; V = W_{t\text{-fusion}} \Big).
\end{equation}
where $Z_i$ denotes latent features at layer $i$. This adds a timestep-aware residual connection that dynamically controls identity influence throughout denoising.
During training, the model receives both (i) the fused identity sequence from Section~\ref{sec:multimodal} and (ii) the time-aware embedding $W_{t\text{-fusion}}$. The full training objective becomes:
\begin{equation}
\begin{aligned}
L_c \;=&\; E_{t,\;z_0,\;y_{\text{user}},\;y_{\text{identity}},\;f,\;\epsilon}\Big[ \big\|\, \epsilon \;- \\ 
&\; \epsilon_{\theta}\!\big(z_t,\, t,\, \psi_{\theta}(y_{\text{user}}, y_{\text{identity}}, f),\, \phi_{\theta}(W_{\text{fusion}}, t)\big) \,\big\|^2 \Big].
\end{aligned}
\end{equation}
extending $L_b$ by incorporating $\phi_{\theta}$ as an additional conditioning path. This allows the model to adaptively apply identity cues over time, ensuring both identity consistency and prompt alignment throughout generation.

\subsection{Adaptive Motion Learning}\label{sec:motion}
Text-to-video diffusion models often struggle to generate realistic motion, especially for complex actions, due to the lack of explicit motion supervision. To minimize reconstruction loss, they tend to avoid motion that risks artifacts. Although prior works~\cite{hu2024animate, xu2024magicanimate, zhu2024champ, ma2024follow} incorporate motion priors or auxiliary inputs, they typically rely on reference sequences or specialized architectures, limiting broad applicability.
To overcome this, we introduce a lightweight, self-supervised motion-weighted loss that guides the model to better reconstruct dynamic regions.

\noindent\textbf{Motion Heatmap Generation.}
Given a training video $x$ with $L$ frames, we compute dense optical flow between consecutive frames using RAFT~\cite{teed2020raft}, then derive per-pixel motion magnitudes to form a heatmap $M \in \mathbb{R}^{L \times H \times W}$. We obtain a binary subject mask $\Omega_{\text{body}}$ using off-the-shelf segmentation models~\cite{chen2018encoder, gruosso2021human}, and isolate subject motion via:
\begin{equation}
M_{\text{body}} = \sigma\big( \Omega_{\text{body}} \odot M \big),
\end{equation}
where $\sigma(\cdot)$ denotes the sigmoid function. Values near 1 indicate high motion, and 0 denote static regions. We downsample $M_{\text{body}}$ to match the latent resolution, yielding $M'_{\text{body}}$, a spatial motion importance mask aligned with the latent $z_t$.

\noindent\textbf{Adaptive Motion-Aware Guidance.}
To emphasize learning on dynamic regions, we reweight the denoising loss as:
\begin{equation}
L_d = \big(1 + \lambda M'_{\text{body}}\big) \odot L_c,
\end{equation}
where $L_c$ is the identity-aware loss from Section~\ref{sec:timeaware}, and $\lambda$ controls the emphasis on motion. This penalizes errors more heavily in high-motion areas.
This simple yet effective weighting encourages the model to learn motion dynamics directly from training data—without requiring motion input at inference. As a result, Proteus-ID generates videos that are noticeably more fluid and temporally coherent—without requiring motion input at inference.

\section{Experiments}

\subsection{Dataset}
\begin{figure*}
	\centering
	\includegraphics[width=0.99\linewidth]{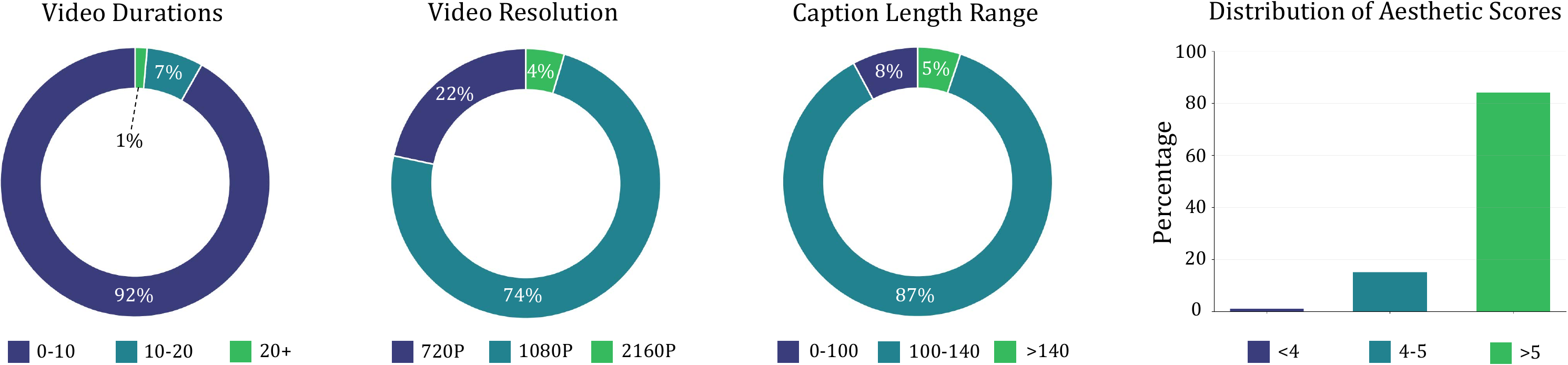}
	\centering
	\caption{\textbf{Video statistics of the dataset.} The dataset comprises diverse video durations and caption lengths, with most videos in 1080P.}
    \label{fig:dataset}
\end{figure*}

We build a high-quality video dataset to support training for video identity customization. Data statistics are shown in Figure~\ref{fig:dataset}. Below, we describe the data construction process.

\noindent{\textbf{Data Curation.}} 
Most existing video identity customization datasets are either proprietary~\cite{zhong2025concat, wei2025echovideo} or limited in scale—e.g., ConsisID~\cite{yuan2024identity}, which contains only ~20K clips. 
To address this gap, we build a large-scale, open-source dataset for the research community.
Following~\cite{yuan2025magictime}, we curated a set of search keywords (e.g., "human", "woman", "man") to collect videos from the internet.
To ensure content quality, we filtered out videos with low engagement (e.g., low view/like counts). We then applied automated filters—PaddleOCR~\cite{zhou2017east, liao2022real}, aesthetic scoring, and motion analysis—following ChronoMagic~\cite{yuan2024chronomagic}, to discard low-quality clips. This yielded a curated dataset of 200K high-quality videos.


\noindent\textbf{Data Filtering.}
To ensure both identity visibility and scene context, we applied multiple post-processing steps. First, we used YOLO-Box~\cite{jocher2022ultralytics} to extract bounding boxes for "face", "head", and "body" in each frame. We then applied YOLO-Pose~\cite{jocher2022ultralytics} to detect key facial landmarks (e.g., eyes, ears, nose) and removed clips with insufficient keypoint coverage. 
To further improve data quality, we remove clips where the "face" bounding box occupies less than 6\% of the frame area.
As a single video may contain multiple individuals, we adopt the identical verification~\cite{yuan2024identity} to assign each person a unique identifier for subsequent training.



\noindent\textbf{Segmentation and Captioning.}
To support adaptive motion learning, we use the highest-confidence bounding box (obtained in previous steps)  for each category to generate individual body masks via SAM2~\cite{ravi2024sam}. SAM2’s tracking signals are further used to refine identity assignments across frames.
For captioning, we adopt the time-aware annotation strategy from~\cite{yuan2024chronomagic} and generate high-quality, temporally aligned video descriptions using Qwen2.5-VL-72B~\cite{bai2025qwen2}.

\begin{figure*}[t]
	\centering
	\includegraphics[width=0.99\linewidth]{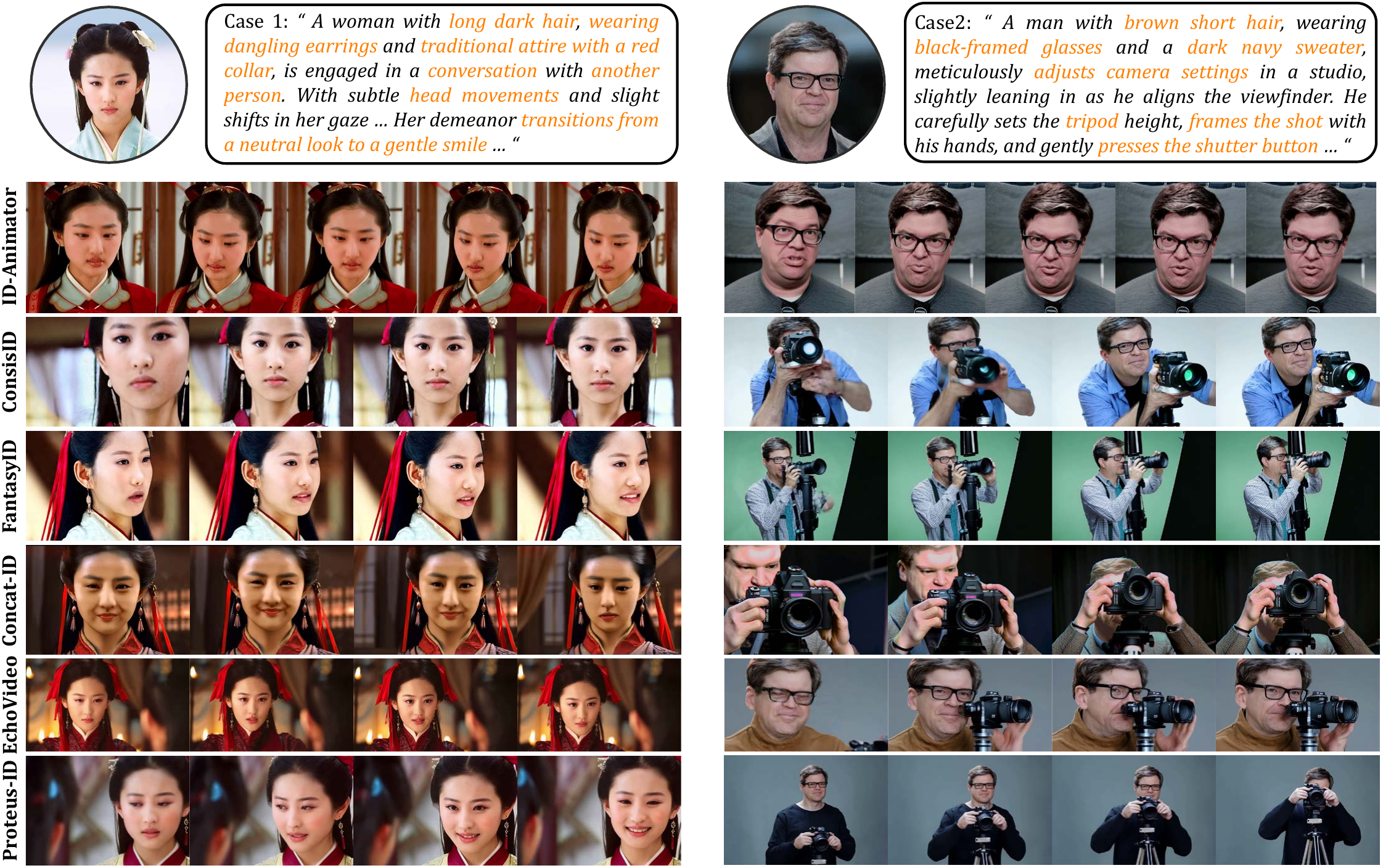}
        \vspace{-2mm}
	\caption{\textbf{Qualitative comparison with state-of-the-art methods.} ID-Animator exhibits poor identity preservation and visual quality. While ConsisID, Fantasy, and Concat-ID improve identity preservation to some extent, they suffer from severe copy-paste artifacts and text misalignment. EchoVideo maintains consistent identity and text alignment but lacks fluid, natural motion. In contrast, our method achieves strong performance in visual quality, identity preservation, text alignment, and coherent motion, substantially outperforming baseline methods.}
        \vspace{-3mm}
	\label{fig:qual}
\end{figure*}

\subsection{Setup}
\noindent\textbf{Benchmark and Metrics.}
In the absence of a standardized benchmark, we curated a set of 150 individuals from public sources, ensuring they were excluded from the training set and selected to represent diverse professions and ethnicities. For each subject, we designed prompts encompassing a range of appearances, expressions, actions, and backgrounds.
Following prior work~\cite{huang2024consistentid, zhang2025fantasyid}, we evaluate performance across four key dimensions:
(1) \textit{Identity Preservation}: We use FaceSim-Arc~\cite{deng2019arcface} and FaceSim-Cur, which compute facial similarity between generated and real images in the ArcFace~\cite{deng2019arcface} and CurricularFace~\cite{huang2020curricularface} embedding spaces.
(2) \textit{Text Alignment}: CLIPScore~\cite{hessel2021clipscore} measures the semantic similarity between each video and its corresponding prompt.
(3) \textit{Visual Quality}: We compute FID~\cite{heusel2017gans} over facial regions, using features extracted by InceptionV3~\cite{szegedy2016rethinking}.
(4) \textit{Motion Amplitude}: We quantify movement within human regions using average dense optical flow~\cite{farneback2003two}.



\noindent\textbf{Baseline Models.}
For a comprehensive comparison, we evaluate five representative open-source baselines:
(1) \textbf{ID-Animator}~\cite{he2024id}, based on AnimateDiff~\cite{guo2023animatediff}, introduces a facial adapter to encode identity features.
(2) \textbf{ConsisID}~\cite{yuan2024identity} injects both global and local facial features to capture low- and high-frequency identity information.
(3) \textbf{FantasyID}~\cite{zhang2025fantasyid} combines 2D and 3D facial embeddings to guide identity-specific video generation.
(4) \textbf{Concat-ID}~\cite{zhong2025concat} uses a VAE to extract identity features, concatenated with video latents along the temporal axis and processed via 3D self-attention.
(5) \textbf{EchoVideo}~\cite{wei2025echovideo} employs a two-stage training strategy, introducing stochasticity in the second stage to selectively incorporate shallow facial cues.


\noindent{\textbf{Implementation Details.}} During training, we set the resolution to \(480 \times 720\) and extract 49 consecutive frames from each video with a stride of 3 to construct the training data.
The Q-Former uses the same structure as described in~\cite{huang2024smartedit}.
TAII consists of 10 units and 32 learnable queries.
We use a batch size of 64, a learning rate of \(3 \times 10^{-6}\), and train for a total of 3.8K steps.
The classifier-free guidance null-text probability is set to 0.1. We adopt AdamW as the optimizer and apply a cosine learning rate scheduler with restarts.
The adaptive motion learning loss weight \( \lambda \) is set to 1.0.
During inference, we employ DPM~\cite{sohl2015deep} with 50 sampling steps and a guidance scale of 6.0.

\subsection{Qualitative Analysis}
\begin{figure*}
	\centering
	\includegraphics[width=0.992\linewidth]{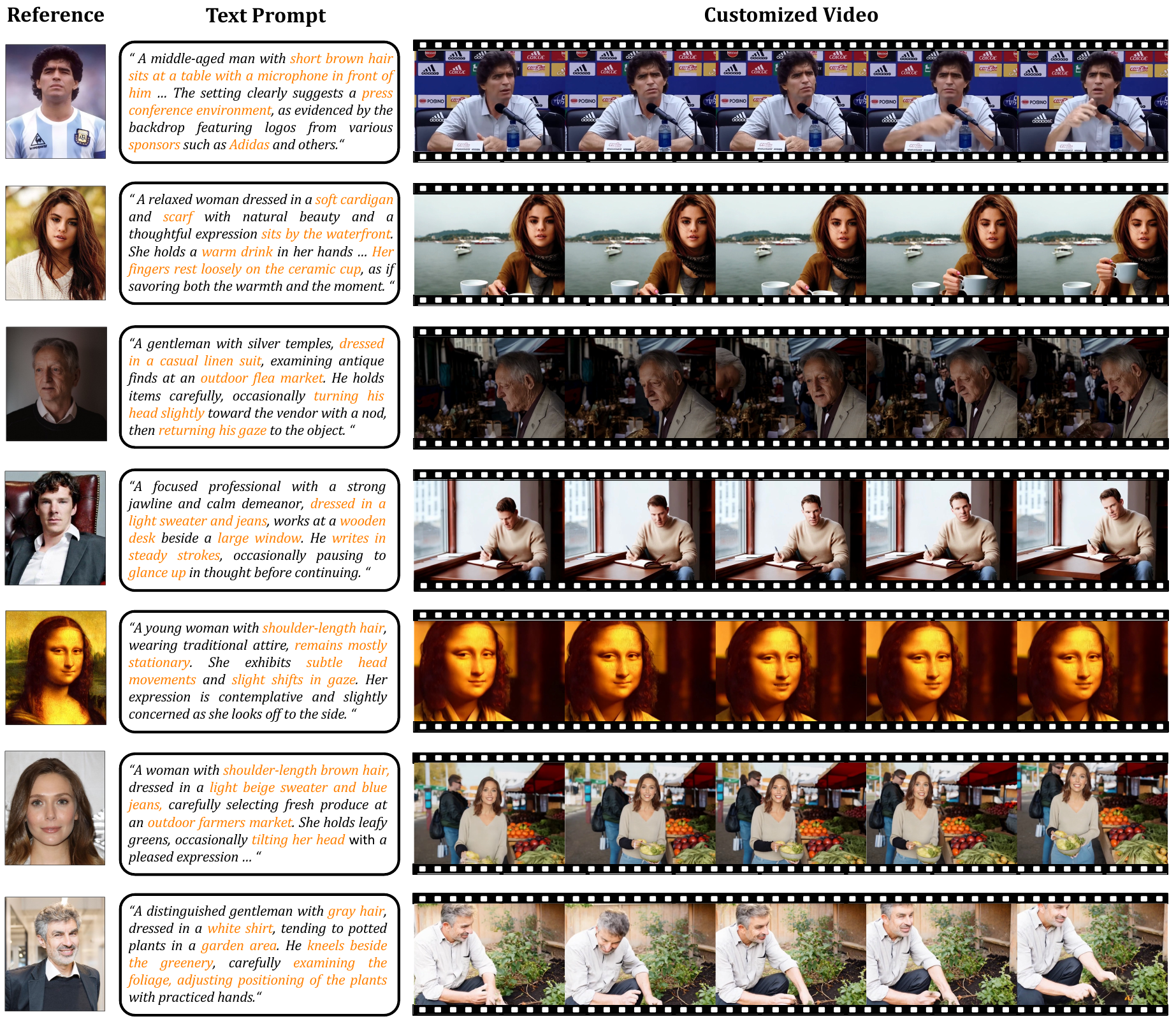}
	\centering
        \vspace{-2mm}
	\caption{\textbf{More qualitative results of Proteus-ID.} Our method generates high-fidelity videos across diverse environmental contexts (indoor/outdoor) and visual styles (photorealistic/artistic), while maintaining precise adherence to both the visual identity from reference images and the attributes specified in text prompts, demonstrating strong practical applicability. \textcolor{orange}{Orange} highlights attributes mentioned in long instructions.}
    \label{fig:appen1}
\end{figure*}

\begin{figure*}
	\centering
	\includegraphics[width=0.97\linewidth]{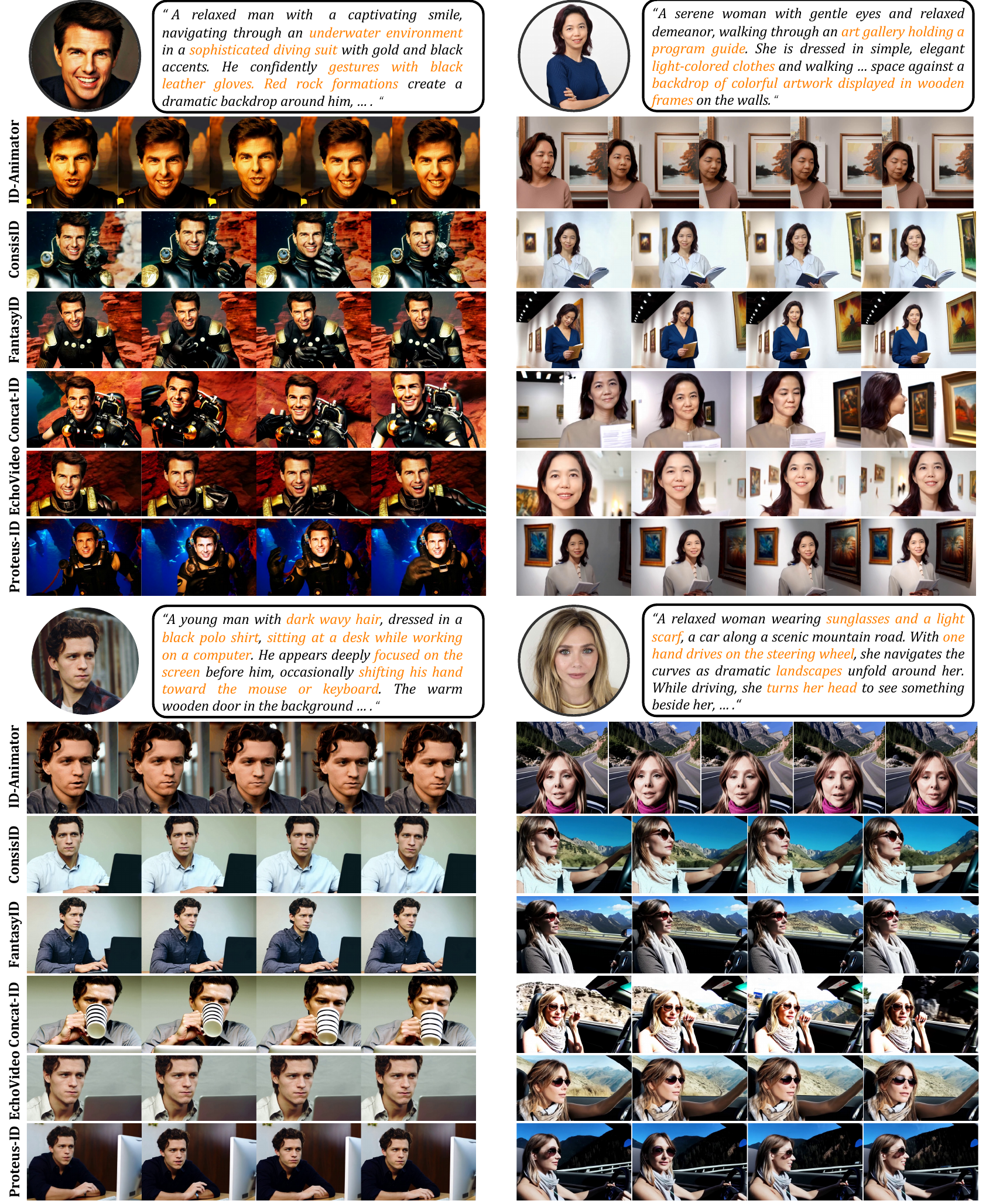}
	\centering
        \vspace{-3mm}
	\caption{\textbf{More qualitative comparisons with state-of-the-art methods.} Proteus-ID demonstrates superior performance across multiple dimensions: identity preservation, visual fidelity, textual relevance, and motion coherence. \textcolor{orange}{Orange} highlights attributes mentioned in long instructions.}
    \label{fig:appen2}
\end{figure*}

This section presents a qualitative comparison between Proteus-ID and baseline methods.
We randomly select images and text prompts of two individuals for qualitative analysis, ensuring that none were included in the training data.
As shown in Figure~\ref{fig:qual}, ID-Animator fails to generate body parts beyond the face and exhibits poor identity fidelity and visual quality.
ConsisID, Fantasy, and Concat-ID partially improve identity preservation but introduce noticeable copy-paste artifacts and exhibit significant text misalignment, as observed in the conversational scenario of case 1.
Although EchoVideo improves text alignment and identity preservation, the absence of motion-aware modeling results in stiff and unnatural motion in cases 1 and 2, compromising visual quality.
In contrast, Proteus-ID achieves superior results in visual quality, identity preservation, text alignment, and coherent motion in challenging scenarios.
Additional qualitative results are presented in Figure~\ref{fig:appen1} and Figure~\ref{fig:appen2}.

\subsection{Quantitative Analysis}
We present a comprehensive quantitative evaluation of all methods, as summarized in Table~\ref{tab:quan}.
Consistent with the results in Figure~\ref{fig:qual}, our method outperforms state-of-the-art baselines across most metrics.
Proteus-ID achieves higher scores in identity preservation (FaceSim-Cur, FaceSim-Arc) and text alignment (CLIPScore) by integrating multimodal features and exploiting their complementary strengths.
In contrast, ConsisID, Fantasy, and Concat-ID rely solely on facial representations, leading to lower CLIPScore performance.
Furthermore, these methods suffer from lower identity preservation due to artifacts caused by semantic conflicts.
In terms of visual quality, ID-Animator achieves the best FID score.
However, this may be due to its tendency to generate more static content.
Please refer to Figure~\ref{fig:qual} for qualitative analysis of visual quality.
Benefiting from explicit motion modeling, our method achieves superior dynamic motion (i.e., Motion Amplitude), enhancing the overall visual experience.
In summary, our method achieves exceptional identity fidelity while effectively preserving text alignment and ensuring coherent motion.

\begin{table*}[t]
\caption{\textbf{Quantitative comparison with state-of-the-art methods.} Proteus-ID achieves well-aligned results across most metrics. $\uparrow$ indicates higher is better; $\downarrow$ denotes lower is preferable. The best result in each column is shown in \textbf{bold}, and the second-best is \underline{underlined}.}
\vspace{-4mm}
\label{tab:quan}
\begin{center}
\scalebox{1.08}{
\footnotesize
\setlength{\tabcolsep}{5mm}
\renewcommand{\arraystretch}{1.1}
\begin{tabular}{c ccccc}
\toprule
Method & FaceSim-Cur $\uparrow$ & FaceSim-Arc $\uparrow$ & FID $\downarrow$ & CLIPScore $\uparrow$ & Motion Amplitude $\uparrow$ \\
\midrule
ID-Animator~\cite{he2024id}      & 0.365 & 0.351 & \textbf{97.253}         & 27.310 & 10.135 \\
ConsisID~\cite{yuan2024identity}          & \underline{0.614} & \underline{0.596} & 126.586                 & 29.075 & 23.491 \\
FantasyID~\cite{zhang2025fantasyid}         & 0.509 & 0.495 & 127.870                 & 28.125 & 15.724 \\
Concat-ID~\cite{zhong2025concat}         & 0.608 & 0.584 & 134.590                 & \underline{29.094} & 26.516 \\
EchoVideo~\cite{wei2025echovideo}         & 0.528 & 0.510 & 149.037                 & 28.100 & \underline{30.148} \\
\textbf{Proteus-ID (Ours)} & \textbf{0.682} & \textbf{0.661} & \underline{117.999} & \textbf{29.235} &  \textbf{30.679} \\
\bottomrule
\end{tabular}
\vspace{-5mm}
} 
\end{center}
\end{table*}

\begin{figure}[t]
	\centering
	\includegraphics[width=0.99\linewidth]{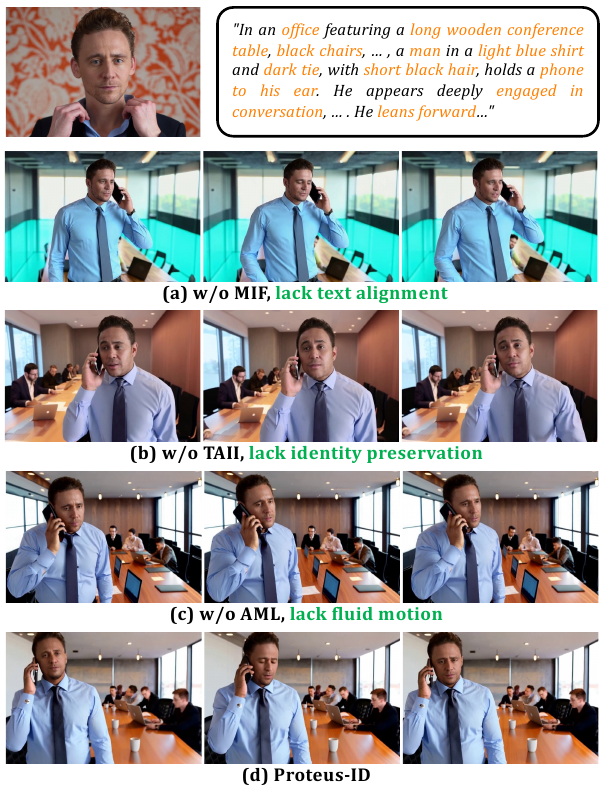}
        \vspace{-3mm}
	\caption{\textbf{Effect of Different Components via Qualitative Analysis.} (a) Removing Multimodal Identity Fusion (MIF) leads to severe text misalignment. (b) Removing Time-aware Identity Injection (TAII) affects identity details. (c)  Removing Adaptive Motion Learning (AML) results in stiff motion.}
        \vspace{-4mm}
	\label{fig:abla}
\end{figure}

\subsection{User Study}
To further assess the effectiveness of our proposed method, we conducted a user study with 30 participants evaluating 120 comparison samples. Participants rated each video on a 1–5 scale across four criteria: (1) Identity Preservation (IP), (2) Text Alignment (TA), (3) Motion Coherence (MC), and (4) Visual Quality (VQ). A minimum response time of one minute was enforced to promote thoughtful and unbiased evaluation. To reduce potential bias, the order of video presentation was randomized for each participant. As shown in Table ~\ref{tab:userstudy}, the results show that Proteus-ID consistently outperforms the baseline across all evaluation criteria, demonstrating superior generation quality in human assessments.

\subsection{Ablation Studies}

\noindent{\textbf{Effect of Multimodal Identity Fusion.}}
To evaluate the impact of the multimodal identity fusion mechanism, we extract only the visual identity features and inject them into the diffusion model. The qualitative results in Figure~\ref{fig:abla} show that removing multimodal identity fusion causes the model to rely solely on visual information, leading to substantial text misalignment.
The quantitative results in Table~\ref{tab:abla} show declines in identity preservation, text alignment, and visual quality, suggesting increased identity-text conflicts.
These results highlight the critical role of multimodal identity fusion in balancing identity preservation and text alignment.

\noindent{\textbf{Effect of Time-aware Identity Injection.}}
We remove time-aware identity injection and directly inject the fused features into the diffusion model.
As illustrated in Figure~\ref{fig:abla}, this variant reduces identity details, thereby impairing the model’s ability to maintain identity consistency.
Table~\ref{tab:abla} additionally demonstrates that this variant results in decreased FSim-Cur, FSim-Arc, FID, and CLIPScore values.
These findings suggest that time-aware identity injection enables adaptive adjustment of guidance strength during denoising, enhancing both identity consistency and text alignment.

\noindent{\textbf{Effect of Adaptive Motion Learning.}}
Removing the adaptive motion learning strategy required the model to simultaneously focus
on both foreground and background elements, with background noise negatively affecting human motion.
Figure~\ref{fig:abla} demonstrates that this change significantly reduces motion amplitude and yields rigid, unnatural movements.
Table~\ref{tab:abla} indicates that adaptive motion learning slightly increases FID and reduces CLIPScore values.
However, this trade-off is acceptable given the overall improvement in visual quality.
\begin{table}[t]
\caption{\textbf{User Study.} Higher scores indicate better performance.}
\vspace{-4mm}
\label{tab:userstudy}
\begin{center}
\scalebox{0.95}{
\small 
\setlength{\tabcolsep}{3mm}
\renewcommand{\arraystretch}{1}
\begin{tabular}{c cccc}
\toprule
Method & IP & TA & MC & VQ \\
\midrule
ID-Animator~\cite{he2024id} & 3.43 & 2.63 & 2.87 & 2.97 \\
ConsisID~\cite{yuan2024identity}    & 3.60 & 3.57 & 3.53 & 3.60 \\
FantasyID~\cite{zhang2025fantasyid}   & 3.47 & 3.47 & 3.60 & 3.57 \\
Concat-ID~\cite{zhong2025concat}   & 3.67 & 3.63 & 3.37 & 3.37 \\
EchoVideo~\cite{wei2025echovideo}   & 3.83 & 3.50 & 3.13 & 3.17 \\
\textbf{Proteus-ID (Ours)} & \textbf{4.03} & \textbf{3.87} & \textbf{3.67} & \textbf{3.63} \\
\bottomrule
\end{tabular}
} 
\end{center}
\end{table}

\begin{table}[t]
\caption{Effect of Multimodal Identity Fusion (MIF), Time-aware Identity Injection (TAII) and Adaptive Motion Learning (AML) by Automatic
Metrics. The best result in each column is shown in \textbf{bold}, and the second-best is \underline{underlined}.}
\vspace{-3mm}
\label{tab:abla}
\centering
\resizebox{\columnwidth}{!}{%
\footnotesize
\setlength{\tabcolsep}{0.1mm}
\renewcommand{\arraystretch}{1.1}
\begin{tabular}{lccccc}
\toprule
Method & FSim-Cur~$\uparrow$ & FSim-Arc~$\uparrow$ & FID~$\downarrow$ & CLIPScore~$\uparrow$ & Motion Amplitude~$\uparrow$ \\
\midrule
w/o MIF & 0.667 & 0.645 & 129.015 & 28.401 & \textbf{30.761} \\
w/o TAII  & 0.661 & 0.639 & 124.071 & 28.704 & 30.557 \\
w/o AML & \textbf{0.685} & \textbf{0.662} & \textbf{116.747} & \textbf{29.610} & 27.439 \\
Proteus-ID & \underline{0.682} & \underline{0.661} & \underline{117.999} & \underline{29.235} &  \underline{30.679} \\
\bottomrule
\end{tabular}
}
\end{table}

\section{Conclusion}
In this paper, we propose Proteus-ID, a framework for identity-consistent and motion-coherent video generation.
The Proteus-ID employs a multimodal identity fusion module to encapsulate both the subject’s appearance and its semantic interpretation.
This design bridges the semantic gap between visual and textual identity cues, significantly mitigating identity-text conflicts.
Moreover, the proposed Proteus-ID framework enhances identity consistency and prompt alignment by integrating timestep embeddings into the fused identity via adaptive transformations, enabling context-aware identity guidance throughout denoising process.
To improve motion realism, we introduce an adaptive motion learning strategy that leverages self-supervised motion signals to guide the model in producing temporally coherent motion without additional input at inference.
Due to the lack of high-quality training data, we construct Proteus-Bench to support and advance research in video identity customization.
Experimental results show that our framework excels at generating high-quality personalized videos and outperforms existing state-of-the-art video identity customization models.

{
\small
\bibliographystyle{ieeenat_fullname}
\bibliography{ProteusID_arxiv}
}

\clearpage
\setcounter{page}{1}

\vspace{-10mm}

\end{document}